\documentclass[journal]{IEEEtran}
\usepackage{amsmath,amsfonts}
\usepackage{algorithmic}
\usepackage{algorithm}
\usepackage{array}
\usepackage[caption=false,font=normalsize,labelfont=sf,textfont=sf]{subfig}
\usepackage{textcomp}
\usepackage{stfloats}
\usepackage{url}
\usepackage{verbatim}
\usepackage{graphicx}
\usepackage{cite}
\usepackage{color}
\usepackage{hyperref}
\usepackage{booktabs} 
\usepackage{tabularray}
\usepackage{multirow}
\usepackage[normalem]{ulem}
\usepackage{xcolor}
\usepackage{ulem}
\usepackage{makecell}
\usepackage{graphicx} 

\newcommand{\toolName}[1]{\textit{ChartKG}}

\newcommand{\majorRevise}[1]{\textcolor{black}{#1}}
\newcommand{\minorRevise}[1]{\textcolor{black}{#1}}

\begin{document}
\title{\toolName{}: A Knowledge-Graph-Based Representation for Chart Images}

\author{
Zhiguang Zhou, Haoxuan Wang, Zhengqing Zhao, Fengling Zheng, Yongheng Wang, Wei Chen and Yong Wang
\thanks{Zhiguang Zhou, and Fengling Zheng are with the School of Media and Design, Hangzhou Dianzi University, Hangzhou, CO 310018 China (e-mail: zhgzhou@hdu.edu.cn; liuyuhua@hdu.edu.cn;).}
\thanks{\emph{(Corresponding authors: Yong Wang)}}
\thanks{Manuscript received April 19, 2021; revised August 16, 2021.}
\thanks{Zhiguang Zhou is with the School of Media and Design, Hangzhou Dianzi University (e-mail: zhgzhou@hdu.edu.cn). Haoxuan Wang, Zhengqing Zhao and Fengling Zheng are with Big Data Visualization and Human Computer Collaborative Intelligent Laboratory, Hangzhou Dianzi University (e-mail: whxxyhf111@gmail.com; zhaozhgqng@gmail.com; Zhengyiida@163.com). Yongheng Wang is with the Research Center of Big Data Intelligence Zhejiang Lab, Hangzhou, China (e-mail: wangyh@zhejianglab.com); Wei Chen is with State Key Lab of CAD\&CG, Zhejiang University (e-mail: chenvis@zju.edu.cn). Yong Wang is with College of Computing and Data Science, Nanyang Technological University (e-mail: yong-wang@ntu.edu.sg).
}
}

\markboth{IEEE TRANSACTIONS ON VISUALIZATION AND COMPUTER GRAPHICS, VOL. xx, NO. x, JUNE 20xx}%
{Shell \MakeLowercase{\textit{et al.}}: A Sample Article Using IEEEtran.cls for IEEE Journals}

\maketitle



\begin{abstract}
    Chart images, such as bar charts, pie charts, and line charts, are
    explosively produced due to the wide usage of data visualizations. Accordingly, knowledge mining from chart images is becoming increasingly important, which can benefit downstream tasks like chart retrieval and knowledge graph completion. However, existing methods for chart knowledge mining mainly focus on converting chart images into raw data and often ignore their
    visual encodings and semantic meanings, which can result in information loss for many downstream tasks. In this paper, we propose \toolName{}, a novel knowledge graph (KG) based representation for chart images, which can model the visual elements in a chart image and semantic relations among them including visual encodings and visual insights in a unified manner.Further, we develop a general framework to convert chart images to the proposed KG-based representation. It integrates a series of image processing techniques to identify visual elements and relations, e.g., CNNs to classify charts, yolov5 and optical character recognition to parse charts, and rule-based methods to construct graphs.
    We present four cases to illustrate how our knowledge-graph-based representation can model the detailed visual elements and semantic relations in charts, and further demonstrate how our approach can benefit downstream applications such as semantic-aware chart retrieval and chart question answering. We also conduct quantitative evaluations to assess the two fundamental building blocks of our chart-to-KG framework, i.e., object recognition and optical character recognition. The results provide support for the usefulness and effectiveness of \toolName{}.

\end{abstract}






\graphicspath{{figs/}{figures/}{pictures/}{images/}{./}} 





\section{Introduction}

\begin{figure*}[ht]
\centering
\includegraphics[width=\linewidth]{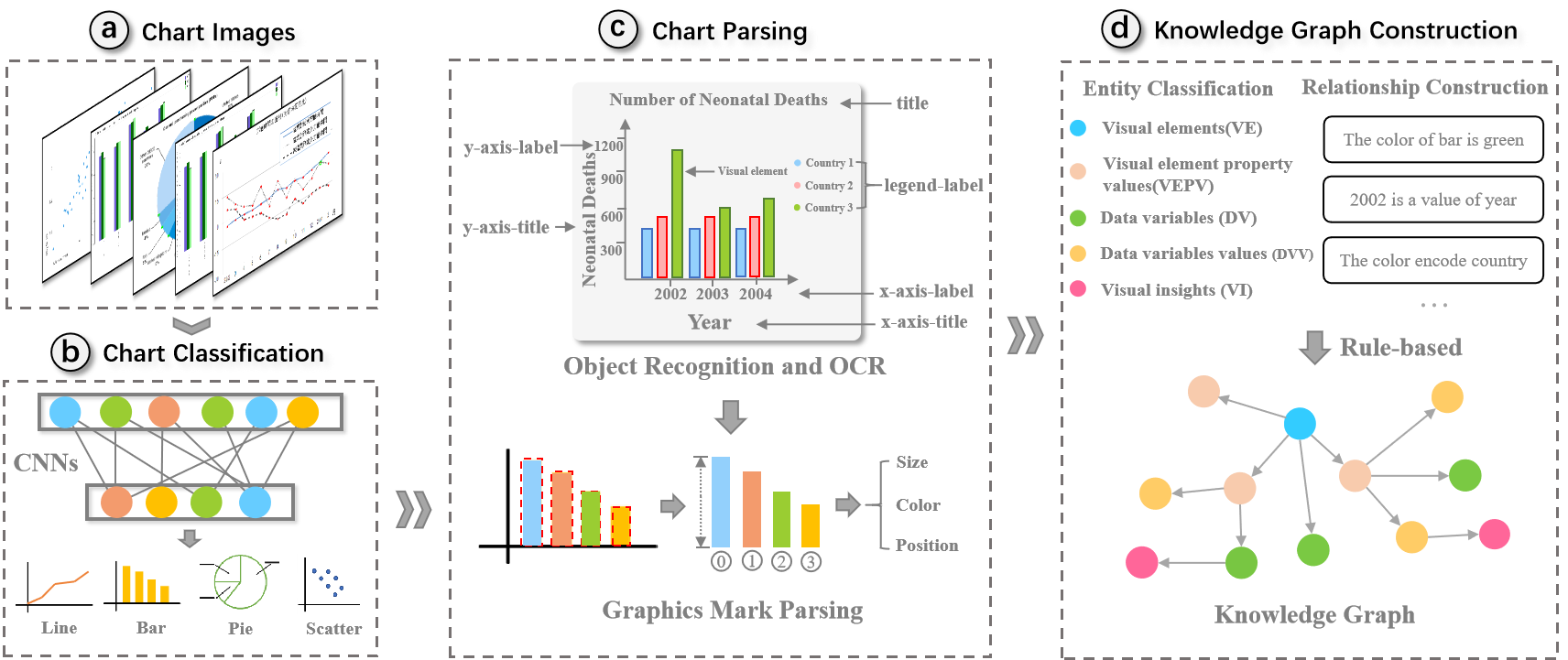}
\caption{ An overview of the framework to convert chart images to the proposed knowledge graph representations.
We first use CNNs to detect input images for chart classification. Then, object recognition and Optical Character Recognition (OCR) are introduced to parsing the charts. A rule-based method is developed to construct the final knowledge graphs for chart images.}
\label{fig1}
\end{figure*}


\IEEEPARstart{D}{ata} visualizations are used almost everywhere and a large amount of chart images (e.g. bar chart, line chart, and pie chart) have been created and accumulated online, where most of them are stored as bitmap images. Accordingly, various benchmark datasets of chart images, like VizML~\cite{hu2019vizml}, VisImages~\cite{deng2022visimages}, and PlotQA~\cite{methani2020plotqa}, have been created. Also, it has become increasingly popular to analyze chart images and conduct knowledge mining from them for various tasks and applications, such as chart retrieval~\cite{li2022structure}, chart redesign~\cite{zhang2021chartmaster}, visual reasoning~\cite{sharma2019chartnet} and chart question answering~\cite{masry2022chartqa}.

Depending on how chart images are represented for subsequent knowledge mining, the existing methods for knowledge mining from chart images can be categorized into two groups: \textit{end-to-end methods}~\cite{sharma2019chartnet, zhang2021chartmaster, ma2018scatternet} and \textit{data-extraction based methods}~\cite{methani2020plotqa, masry2022chartqa, kanthara2022chart}. The end-to-end methods often leverage deep neural networks to directly represent the hidden knowledge in the original chart images for the corresponding tasks. For example, ScatterNet~\cite{ma2018scatternet} leverages deep learning models to capture the subjective similarity of scatterplots. However, such
deep learning-based 
methods are often designed for one specific task and cannot be generalized to other purposes. Also, these methods usually work like a blackbox and lack explainability~\cite{sharma2019chartnet}. The data-extraction-based methods often 
conduct reverse-engineering and convert the original chart images to the extracted raw data of charts, and then all the subsequent analyses will be built upon the extracted raw data. 
For instance, ReVision~\cite{savva2011revision} utilizes image processing techniques to identify graphical marks and infer the underlying data, which is further used for the redesign of charts.
\majorRevise{But these methods focus on extracting the data encoded in charts, and do not explicitly provide a generic and easy-to-use representation to preserve the critical information of charts such as data, visual elements and the corresponding relationships among them (e.g., the height of a bar in a bar chart encodes the math score of a specific student), which, however, is critical for many downstream chart image analysis tasks like automated infographics design~\cite{chen2019towards}.}
In summary, a unified, expressive, and explainable representation for chart images that can facilitate the knowledge mining of chart images is still missing.

Inspired by the recent research in knowledge graphs~\cite{li2021kg4vis}, we aim to explore the possibility of representing chart images as knowledge graphs to facilitate downstream knowledge-mining tasks of chart images. However, it is a non-trivial task and the challenges originate from two major aspects: \textit{unified representations} and \textit{automated extraction of entity and relationship}. First, there are various charts with different visual elements and visual encodings. It is difficult to represent all the necessary information of different chart images in a unified, expressive, and explainable manner. Second, chart images consist of different visual elements and encode data with different visual channels. It remains unclear on how to automatically extract visual elements as well as their relations with a good accuracy.

To tackle the above challenges, we develop \toolName{}, a novel knowledge-graph-based representation for chart images to facilitate unified and explainable knowledge mining from chart images. Specifically, by investigating and summarizing common visual elements of charts and the semantic relations among them, we categorize the involved entities of charts into five types: visual elements (e.g., bars in a bar chart), visual element property values (e.g., the height of a bar), data variables, the corresponding data variable values and visual insights (e.g., trend). We further group the semantic relations among them into four types: visual property correspondences (i.e., the relationship between visual elements and their visual property values), data variable correspondences (i.e., the correspondence between data variables and their values), visual encoding mappings (i.e., how a data variable is encoded as a specific visual property) and visual insight correspondence 
(i.e., the correspondence between variable(s) and visual insight(s)). The entities and relations collectively form a knowledge graph (KG), which is a unified and expressive representation for various charts.
Then, to implement our KG-based representation, we develop a general framework to extract the entities and relations from chart images, which incorporate state-of-the-art image processing techniques such as convolutional neural networks~\cite{krizhevsky2017imagenet}, YOLOv5~\cite{jocher2021ultralytics}, optical character recognition~\cite{smith2007overview} and rule-based methods. 

It can automatically convert input chart images into our proposed KG-based representations. We present case studies and two example application scenarios such as semantic-aware chart retrieval and chart question answering to showcase the usefulness and effectiveness of our approach. Also, we conduct quantitative evaluations to demonstrate the effectiveness of the two major building blocks of the chart-to-KG conversion framework of \toolName{}, i.e., object recognition and optical character recognition.

In summary, the main contributions of our work can be summarized as follows:

\begin{itemize}
\item{A novel knowledge-graph-based representation for chart images is proposed to characterize the visual elements and their relationships in an expressive and unified manner.}
\item{A unified framework is designed to automatically convert chart images to the proposed knowledge-graph-based representation.}
\item{Case studies, example applications, and quantitative evaluations demonstrate the usefulness and effectiveness of our approach in representing chart images, which can benefit downstream tasks such as semantic-aware chart retrieval and chart question answering.}
\end{itemize}

\section{RELATED WORK}
\label{section2}

\subsection{Chart Information Extraction}
There have been a series of research studies on extracting various information, such as the underlying data and detailed visual encodings, from charts. Such extracted information will be further used for visualization re-designing or indexing. For example, Jung et al.~\cite{jung2017chartsense} proposed an interactive chart data extraction tool ChartSense, which extracts the underlying data from chart images based on a deep learning-based classifier and a semi-automatic interactive extraction algorithm. Méndez et al.~\cite{mendez2016ivolver} presented a web-based tool iVoLVER, which requires a relatively large number of interactions to accurately extract data from chart images and reconstruct the representation of data. ReVision~\cite{savva2011revision} is an automatic approach that classifies chart types and extracts data from chart images, which can extract data from bar charts and pie charts. Choi et al.~\cite{choi2019visualizing} developed a Google Chrome extension to automatically extract data from charts in web pages and help people with visual impairment understand the web content. Harper and Agrawala ~\cite{10.1145/2642918.2647411} presented a method to deconstruct D3-based visualizations and extract data, marks and the mappings between them. 
Similarly, Poco et al.~\cite{poco2017reverse} utilized inferred text elements to recover visual encodings and data information from a chart image, to support indexing, searching, or retargeting of the input visualization. \majorRevise{Masson et al.\cite{masson2023chartdetective} developed an interactive system capable of accurately extracting data from charts in the SVG format. Chen et al.\cite{chen2019towards} focused on the automatic design of timeline infographics by extracting visual elements from existing timeline infographics. Chartreuse \cite{cui2021mixed} was proposed to enable the reuse of bar charts through the integration of the re-editing of visual elements. Furthermore, Mystique \cite{chen2023mystique} was designed to deconstruct and repurpose SVG images via user interactions, enhancing the flexible utilization of infographic elements.}

Our work is inspired by the above existing studies in terms of chart information extraction. However, almost all the existing work stores the extracted data and visual encodings in an ad-hoc manner, which affects the effective usage of such extracted chart information for downstream tasks like chart retrieval, visual question answering and chart summarization. Our work fills such a gap by presenting a unified knowledge-graph-based representation for chart images. Also, our approach further extracts data insights presented in the data visualization, as prior research~\cite{lundgard2021accessible} has shown that it is necessary to present semantically meaningful visualization content to enhance the accessibility of data visualizations.

\subsection{Knowledge Graph Construction for Images}
Recent years have witnessed the resurgence of knowledge engineering featured by the fast growth of knowledge graphs~\cite{9961954}. 
A knowledge graph (KG) is essentially a semantic network that models the semantic relationships among different entities and has been applied to modeling different types of data including images.
Specifically, modeling images as knowledge graphs has been used in a wide range of real-world applications, including image understanding~\cite{8953301,9492217}, visual retrieval~\cite{9194506} and visual question answering~\cite{9199272}. 
For example, various methods have been proposed to generate scene graphs (SG) for natural images, 
which serves as an abstraction of objects and their complex relationships within scene image \cite{Lou_2022_CVPR}. It not only provides fine-grained visual cues for low-level recognition tasks, but has further proven their applications on numerous high-level visual reasoning tasks, such as visual question answering (VQA)~\cite{6818956,zheng2019reasoning}, image captioning~\cite{Chen_2020_CVPR, 5487377} and scene synthesis~\cite{jiang2018configurable, johnson2018image}. 
Besides natural images, semantic knowledge graphs have also been constructed to model the semantic relationship and domain knowledge of remote sensing images~\cite{9553667,sun2022remote}.
In summary, knowledge graphs have been used to represent the semantic information of various images.
Inspired by the powerful capability of knowledge graphs in representing the semantic information of various images, we propose leveraging knowledge graphs to model the semantic information of chart images.

\subsection{Chart Retrieval}
Charts have emerged as valuable search targets, especially when they contain data not easily obtainable through alternative sources. With a plethora of chart data at hand, the need to retrieve charts tailored to a user's specific requirements becomes paramount. At the core of chart-based queries lies the primary data retrieval question, a concern that has been explored in earlier research~\cite{huang2007system}. This question can often be addressed by executing essential operations on the extracted chart data. Some studies have leveraged template-based questions, offering a structured way to specify data sources and variables~\cite{siegel2016figureseer}. Existing methods for visual chart retrieval primarily fall into two categories: task-oriented approaches, primarily focused on matching queries to charts, and keyword-based techniques~\cite{chen2015diagramflyer, 7888968}. The latter category involves matching queries against specific textual roles or chart-style properties~\cite{chen2015diagramflyer, hoque2019searching}. Additionally, certain methods prefer more elaborate queries, employing automated entity detection techniques within queries, and comparing them to comprehensive textual descriptions of the charts. These identified entities are then matched with the content stored within the index, thereby enhancing matching accuracy~\cite{li2015novel}. To improve query completeness, some approaches consider query expansion by including synonyms of query keywords~\cite{chen2015diagramflyer, li2015novel}. However, these methods are often constrained to basic keyword searches, which may not adequately capture complex semantic aspects such as trends. Our method overcomes this limitation by effectively representing the fundamental elements in charts and their relationships, while also embedding visual insights akin to trends. Leveraging ChartKG, we enable more sophisticated queries based on entities and relationships, encompassing aspects like visual encoding and visual insight.

\subsection{Visual Question Answering}
The Visual Question Answering (VQA) task is designed to produce informative responses to natural language questions about specific images. It represents a convergence of computer vision and natural language processing (NLP), holding significant promise for advancing interdisciplinary research. Currently, two prominent methodological models dominate the VQA landscape.

The first model is the classification-based visual QA model, which leverages encoders to represent both the query and the image. It employs attention mechanisms to seamlessly merge the distinctive features of the query and image before performing classification. Kafle et al.\cite{kafle2018dvqa} have contributed substantially to this approach. They have introduced robust baselines, including an end-to-end neural network and a dynamic local dictionary model. Their work not only automates the extraction of numerical and semantic information from bar graphs but also incorporates a chart QA algorithm that intelligently combines question and image features. This intelligent fusion facilitates the aggregation of learned embeddings to effectively answer posed questions\cite{kafle2020answering}. Additionally, architectures focused on chart element localization and QA encoding for chart elements~\cite{singh2020stl, chaudhry2020leaf} have offered valuable insights into improving QA methods. However, it's crucial to note that these approaches are often tailored to specific local problems. Another distinct approach is the table QA method. It either assumes the presence of a data table for the image~\cite{kim2020answering, masry2021integrating} or employs visual techniques to extract the data table directly from the image~\cite{methani2020plotqa, masry2022chartqa}. While these methods address certain aspects of VQA, they frequently lack scalability and may be confined to particular tasks. Conversely, by relying on ChartKG, various types of questions can be easily expanded and answered through entity and relationship matching. At the same time, the ChartKG can provide more contextual information to help the deep learning model comprehend questions and answers. Therefore, knowledge graphs-based VQA methods possess scalability and generalization ability, making them applicable to a wider range of scenarios and tasks.

\section{KG-based Representation for Chart Images}
\label{section3}
\majorRevise{Chart exists in various forms, such as bitmap images, SVG, and program specifications~\cite{chen2023state}. In this paper, we focus on chart images, i.e., charts in the format of bitmap images, since they are widely seen and charts in other formats like SVG and program specifications can be easily converted to bitmap images.} Therefore, we create a knowledge graph to depict the relationships among the entities within the bitmap-image-based charts.

According to the previous work~\cite{saket2018task}, the line chart, bar chart, pie chart and scatter plot are the widely used and basic chart types. We select these four types of charts as representatives to achieve a unified expression of the knowledge graph, which is convenient for future extension to other chart types. Since entities and relationships form the core components of our knowledge graph, we will elaborate on our knowledge-graph-based representation by defining entities and relationships.

\minorRevise{
{\bf{Entities.}} The initial step in constructing our knowledge graph is to define the entity type. Charts consist primarily of or a combination of the following elements~\cite{10.1145/2740908.2742831}: title, x-axis title, y-axis title, legend title, x-axis label, y-axis label, legend label, and graphical mark. Based on these elements we constructed five types of entities.
}
\minorRevise{
\begin{itemize}
\item{Visual elements (VE), are the key components of charts, such as graphical marks (bar, line, and so on) and axis.}
\item{Visual element property values (VEPV), such as bar height or line color, are typically encoded with different data values and are therefore considered entities in the knowledge graph.}
\item{Data variables (DV), including x-axis-title, y-axis-title, and legend-title, which contain crucial semantic information pertaining to data values. }
\item{Data variable values (DVV), including x-axis-label, y-axis-label, and legend-label, which convey specific data values for both categorical and continuous data, are also essential entities for proper comprehension of the chart.}
\item{Visual insights (VI). we consider the easily perceptible visual insights within charts as a distinct entity type. By grouping these elements into separate entities, we can more effectively model the relationships between them in our knowledge graph.}
\end{itemize}
}

\begin{figure}[ht]
    \begin{center}
    \includegraphics[width=\linewidth]{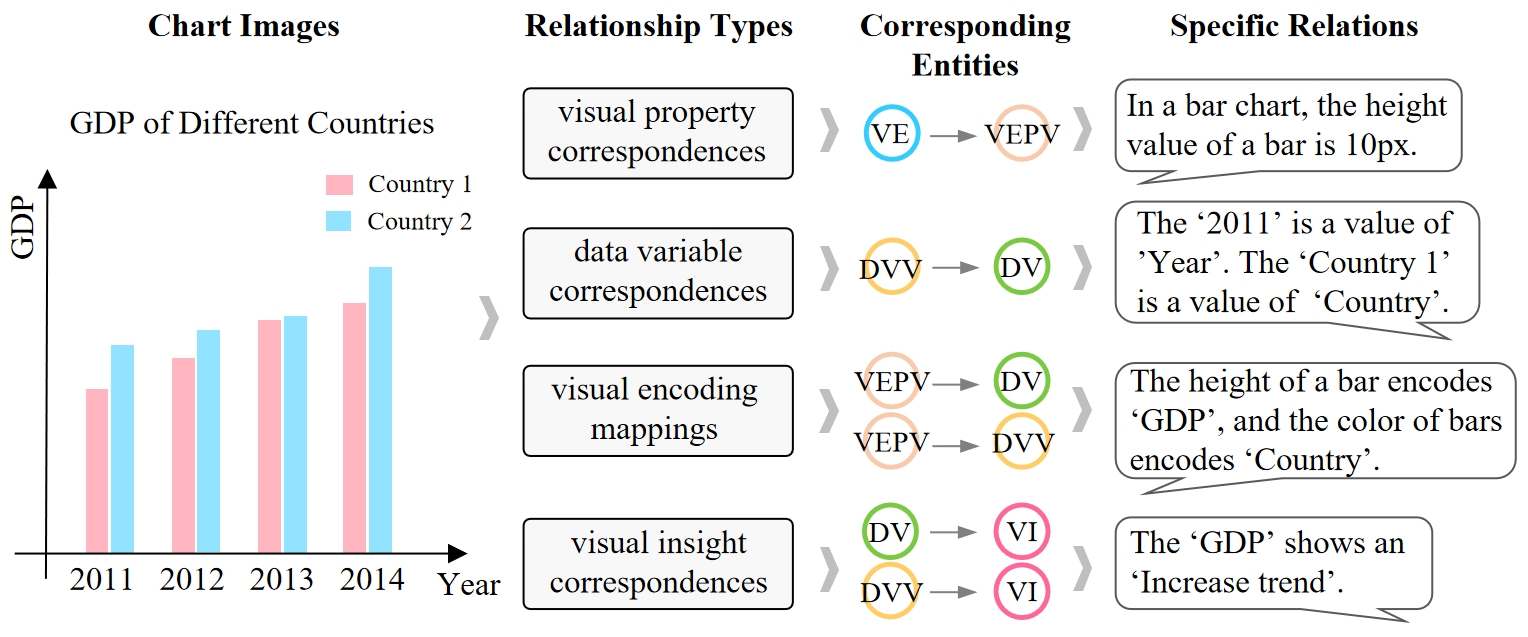}
    \caption{An overview of the relationship types and their corresponding specific relationships. The VE, VEPV, DV, DVV and VI respectively denote visual element, visual element property value, data variable, data variable value and visual insight.}
    \label{table1}
    \end{center}
\end{figure}

\minorRevise{{\bf{Relationships.}} After introducing five classes of entities, we further define four classes of semantic relationships between different entities within a chart: visual property correspondences, data variable correspondences, visual encoding mappings and visual insight correspondences. A comprehensive list of these relations can be found in Figure ~\ref{table1}.}
\minorRevise{
\begin{itemize}
\item{Visual property correspondences. This type of relationship aims to represent the property values of the visual elements. It connects visual elements to visual element property values (VE $\rightarrow$ VEPV), for example, ``the color of a bar is blue".}
\item{Data variable correspondences. It focuses on specific instances of data variables, for example, ``England" and ``America" can be the value of the ``Country" variable. We connect data variable values to data variables (DVV $\rightarrow$ DV), representing the relationship between variables and values, such as ``2011 is a value of Year".}
\item{Visual encoding mappings. The visual encoding mappings express the semantic information conveyed by the underlying data and model the mapping from visual element property values to data variables or data variable values (VEPV $\rightarrow$ DV), for example, ``the color of a bar is blue representing England.".}
\item{Visual insight correspondence. It is employed to represent complete visual insights, which typically describe the underlying characteristics between different data variables and data variable values. We establish relationships between the existing visual insights and corresponding data variables/data variable values (DV$\rightarrow$VI and DVV$\rightarrow$VI) to completely represent the quick insights. QuickInsight~\cite{ding2019quickinsights} presents 12 data insights that are easily and rapidly perceptible across various visualization charts. Using these insights as a reference, we have embedded the 12 visual insights into our knowledge graph, aiding machines to quickly retrieve charts that match insights.
}
\end{itemize}
}

Once all entities and relations are defined, we extract triplets from existing chart images. These triplets serve as edges connecting individual entities and thereby constructing a knowledge graph representation for a chart. Our knowledge graph provides a comprehensible representation of the relations between chart elements and data information by assigning them to different nodes. Our KG-based representation enables quick and effective chart downstream tasks such as chart retrieval and chart question answering.

\section{Chart-to-KG Conversion Framework}
\label{section4}
Chart-to-KG is a framework to generate knowledge graphs from chart images without any user interaction. Determining the chart type and subsequently extracting chart elements based on the specific chart type is a fundamental step. Different types of charts contain distinct chart elements, making it essential to identify the chart type to accurately extract these elements. So the framework mainly consists of three modules: chart classification, chart parsing, and knowledge graph construction as shown in Figure~\ref{fig1}. Firstly, Convolutional Neural Networks (CNNs) are applied to classify an input chart image to identify the chart type (Section ~\ref{section4.2}). Then we introduce different object recognition methods for different chart types (Section ~\ref{section4.3}), to ensure the scalability of the framework. Finally, we provide a series of rules to construct a knowledge graph of chart elements (Section ~\ref{section4.4}).

\subsection{Chart Classification}
\label{section4.2}
We categorize the input chart images into four groups: bar chart, line chart, pie chart and scatter plot by ResNet50~\cite{He_2016_CVPR}. The domain of chart classification has also witnessed the extensive utilization of deep learning techniques~\cite{simonyan2014very, huang2017densely, 9984953}. Among these options, ResNet50 stands out as a suitable choice for chart classification due to its relatively low number of parameter counts and high accuracy~\cite{ye2022visatlas}. We choose ResNet50 and employ a pre-trained ResNet model on the Imagenet dataset~\cite{russakovsky2015imagenet}. We fine-tuned the model on our corpus using the Adam optimizer with a learning rate set to 0.0005. The input chart image will be divided into four types by the trained model. After testing, the model achieved an accuracy of 87.2\% on our dataset. 

\subsection{Chart Parsing}
\label{section4.3}
To extract entities and relationships from a chart, it is essential to analyze the various elements used to convey information in the chart. In this section, we will outline the three fundamental steps of chart parsing: object recognition, optical character recognition, and graphics mark parsing.
\subsubsection{Object recognition}

The initial step to extract entities from charts is to extract all its elements. We summarized eight types of elements from common standard charts (Section ~\ref{section3}) since different types of charts contain various visual elements. Although their visual forms are different, most common standard charts consist of some or all of these elements. Many object detection models are proposed to detect the bounding boxes and recognize the object class in natural images. YOLOv5~\cite{jocher2021ultralytics} is a well-known object detection model with remarkable accuracies and scalabilities. Also, it can be easily extended to detect elements within charts. Therefore, we use YOLOv5 to detect the categories and bounding boxes of the chart elements. \majorRevise{Due to the different types of elements and features in various charts, we adopt the approach of training an object detection model for each chart type to improve the scalability of our method. This approach allows us to add new chart types more efficiently. When adding a new chart type, we only need to further train the model on the data specific to these new types, without the need of retraining the model on the entire dataset containing all chart types.
The corpus required for training the model includes annotations of the bounding boxes of chart elements, as well as their corresponding labels under that chart type. 
The corpus will be introduced
in Section~\ref{section4.1}. Once the training of the model is complete, we can use it to predict the labels and bounding boxes of elements in the new charts. The final YOLOv5 model returns one bounding box for each element within the charts, with possible labels and the bounding box.
} Furthermore, we observed that the accuracy of recognizing text and symbols within legends was relatively low. Therefore, we developed a specialized legend recognition model to enhance the accuracy of identifying color markers and text within legends. Once we obtain the chart type through the classification model, we can easily determine the element categories and bounding boxes using the corresponding element extraction and legend extraction models.

\subsubsection{Optical character recognition}
 Texts within a chart are crucial for conveying the meaning of the underlying data semantics. For example, the title, x-axis label, and y-axis label provide detailed context about the data being presented. It is imperative to accurately identify the textual content in order to extract meaningful semantic information from a chart. To extract specific textual information from chart images, optical character recognition (OCR) technology is used in our framework. Numerous techniques have been proposed in the literature for extracting texts from charts~\cite{smith2007overview,shi2016end,baek2019character}. Our study employs Tesseract~\cite{smith2007overview}  an open-source OCR engine developed by Google, to enable the retrieval of textual information from identified bounding boxes of texts, which the object recognition model detects. \majorRevise{Given the robust text recognition capabilities of the pre-trained Tesseract module, we directly utilize it to extract text content from text bounding boxes at this stage. First, we crop the images based on the bounding boxes and magnify them threefold to enhance character recognition accuracy. The resulting partial images are then input into the Tesseract module to obtain the final text content. By combining the text roles obtained from the Object Recognition phase, we acquire complete textual data with both their role and the context, along with the confidence score.} In addition, our experiments with the pre-trained Tesseract module on our dataset demonstrate its strong performance (Section ~\ref{section7.2}).

\subsubsection{Graphics mark parsing}
\label{markparseing}
For standard charts, graphic marks represent the main encoding method for data, and the major differences between different types of charts rely on the use of graphical marks. Hence, it is necessary to analyze the graphical marks specific to each type of chart. According to the prior study~\cite{bertin1983semiology}, position, color, and size are the most commonly used visual channels. Therefore, we will specifically introduce how to parse the graphics mark focus on size, color, and position in bar charts, line charts, pie charts and scatter plots.

{\bf{Bar chart.}} The graphics marks in bar charts are bars. We will extract the size, color value, and position index of the bars. The object detection model is responsible for identifying bars within bar charts and determining the coordinates of their bounding boxes, specifically the top-left and bottom-right coordinates. In the case of vertical bar charts, the height of each bar serves as a size indicator. This height can be computed by subtracting the minimum y-value from the maximum y-value found within the predicted bounding box. To extract positional information, we utilize the bounding boxes. By arranging the bars based on their x-coordinates, we establish a position index. This index proves valuable for recognizing data patterns and trends. As for color, we gauge the proportions of various colors within the bar's bounding box pixels. Subsequently, we determine the predominant color within the region, effectively mitigating any influence from background colors.

{\bf{Line chart.}} For line charts, it's essential to obtain the coordinates of the start and end points of each line segment. A line chart uses the vertical coordinate of each point to symbolize the corresponding value. To extract information about the start and end points and the color of the polyline, the input image is transformed from the RGB color space to the HSV color space which helps in handling color information more effectively, and then the edge detection algorithm is applied to identify edges in the image. Subsequently, we can obtain the position of the start and end points and the polyline color by calculating the coordinates and color values of edge pixels. We apply a similar method as for the bar chart to get the position index of each point.

{\bf{Pie chart.}} For pie charts, the area (and the angle) of each pie slice encodes the underlying data, and different colors are assigned to each. \majorRevise{To parse the angle and color of the pie chart, we use YOLOv5 to detect the region composed of all pie slices in the pie chart, and crop out this region to generate a histogram of pixel colors.} Additionally, we calculate the angle of each pie slice by determining the ratio of pixel counts of each color with respect to the total number of pixels.


{\bf{Scatter plot.}} For scatter plots, the x-axis coordinates, y-axis coordinates and colors often encode important information. We extract the coordinates information of each point through the coordinates of the bounding box. We further analyze the pixel color distribution in the bounding box and extract the color value of the point as the method in pie charts.

\minorRevise{Since \toolName{} primarily focuses on the visual semantic information of charts, raw data extraction was not performed after parsing the graphics mask. In scenarios where raw data is required, the visual element property values in the KG can be further converted into specific data.}

\subsection{Knowledge Graph Construction}
\label{section4.4}
After completing the chart parsing process, we perform entity classification and relationship construction based on the extracted content, thus building the final graph.

{\bf{Entity classification.}} 
We initially categorize the parsed chart elements into five distinct entity types. According to the entity definitions outlined in section ~\ref{section3}, the graphic marks extracted from the charts are treated as multiple entities, denoted as VE, with each mark's property values such as color and size categorized under the entity type VEPV. As for DV and DVV, the title of chart components typically describes variables, while the labels of the chart components are generally used to represent specific variable values. Therefore, we classify the textual content in the chart into two categories, DV and DVV, based on their text roles. Finally, visual insights encapsulate vital semantic information represented within charts, challenging to directly glean from raw data. We collate the property values of graphical marks into data tuples (e.g., $\left \langle height, index, color \right \rangle$). \majorRevise{Furthermore, we converted the color and index from the data tuples into rows and columns, and used the height as the data value. Then we transform them into a table, which is the input format required by QuickInsights.} Then we extract the insights within a chart and treat each extracted visual insight as an individual entity building upon QuickInsights~\cite{ding2019quickinsights}.

\majorRevise{{\bf{Relationship Construction.}} After classifying entities, the relationships including visual property correspondences,
data variable correspondences, visual encoding mappings, and visual insight correspondences will be constructed using rule-based methods as follows:
\begin{itemize}
\item{Visual property correspondences. The visual property correspondences describe the properties of graphical marks. When we extract the visual elements, the property values of visual elements will be parsed, as shown in Section~\ref{markparseing}. Then the link between the visual element and its property values will be constructed.  For example, we will link a bar with the value of the bar height.}
\item{Data variable correspondences.
The data variable correspondences focus on
specific instances of data variables, for example, “England”
and “America” can be the value of the “Country” variable. After classifying the entities, we obtained DVs and DVVs associated with the chart element labels. Based on the structure of the chart elements corresponding to the data variables and their values within the charts, we established rules, such as the label being an instance of the title. Typically, in charts with axes, the variable values represented by x-labels are instances of the variables represented by x-titles. For example, ``2011" is an instance of ``Year".
}
\item{Visual encoding mappings. The visual encoding mappings represent the semantic information of the underlying data, modeling the relationship between visual element properties and data variables or their values. To establish the visual encoding mapping relationship, we link the VEPV to the DVV and the VEPV to the DV, according to the similarities such as distance similarity, color similarity and so on. 
For example, for bar charts, we calculate the distance between each x-axis label and the bar by the left corner of their bounding boxes, and we designate the text of the x-axis label located closest to the bar as the position index of the bar. 
Furthermore, the relationship between the bar and legend text is established by determining the similarity between their respective colors.} 
\item{Visual insight correspondences. The visual insight correspondence relationships are utilized to represent complete visual insights, which typically describe the underlying characteristics between different data variables and their values. When we extract a visual insight and treat it as an entity, we establish a link between the data variables involved in the computation of this visual insight and the visual insight itself. For example, after extracting the linear correlation between the x-axis title and the y-axis title, the links ``x-axis title $\rightarrow$ linear correlation" and ``y-axis title  $\rightarrow$  linear correlation" will be constructed. 
}
\end{itemize}
}

\section{Quantitative Evaluation of Chart-to-KG Conversion Framework}
\label{section7}
To evaluate the effectiveness of our framework in constructing the KG-based representation for charts, we conducted quantitative assessments of its key components, object recognition and optical character recognition. Section ~\ref{section4.1} describes the datasets used for these evaluations.

\subsection{Corpus}
\label{section4.1}
Since our framework encompasses chart classification, object recognition, and optical character recognition, it is imperative that we have a sufficient amount of training data to train our models and employ testing data to validate the effectiveness of our framework. Due to the limited computation resources, we obtained a subset corpus from the PlotQA~\cite{methani2020plotqa} with rich chart elements annotation information, which includes three chart types: bar charts, line charts, and dot plots. Since the difference between a dot plot and a line plot primarily lies in whether the dots are connected or not, we replaced the dot plots with pie charts and scatter plots which are used more widely. Using the matplotlib package, we generated a batch of charts with corresponding annotation information automatically. Our final corpus consists of four types of charts, each with a quantity of up to ten thousand. To be consistent with the procedure used in PlotQA, we randomly selected 70\% of each chart type for training, 15\% of each chart type for validation, and 15\% of each chart type for testing. 
Then the corpus will be used to evaluate the performance of the object recognition and OCR.

\subsection{Object Recognition Evaluation}
\label{section7.1}
{\bf{Metrics.}} We assessed the performance of object recognition using several key evaluation metrics, including Mean Average Precision (mAP)~\cite{everingham2010pascal}, Precision, and Recall. These metrics are fundamental in evaluating the effectiveness of object recognition models. Mean Average Precision is a widely used metric that provides a comprehensive measure of object recognition performance. It considers the precision-recall curve and calculates the average precision across different levels of confidence thresholds. Specifically, we calculated mAP at two different Intersections over Union (IOU) settings, where IOU measures the degree of overlap between predicted and ground truth bounding boxes. The first setting, mAP at IOU 0.5, focuses on evaluating the model's ability to accurately detect objects when there is at least a 50\% overlap between the predicted and ground truth bounding boxes. The second setting, mAP from IOU 0.5 to 0.95, provides a broader perspective, considering the model's performance across a range of IOU thresholds. Precision evaluates recognition accuracy, representing the ratio of correctly predicted objects to the total number of predicted objects. In contrast, recall measures the model's capacity to identify all instances of an object class by calculating the ratio of correctly predicted objects to the total ground truth objects of that class. By employing these evaluation metrics, we were able to comprehensively analyze and report the performance of our object recognition method, providing valuable insights into its precision, recall, and across different IOU thresholds.

{\bf{Results.}} Table ~\ref{table4} shows the evaluation results, which indicate that our framework performs well in terms of the metrics. Across various object categories, the majority of them exhibit excellent recognition rates, with mAP50, Precision, and Recall scores consistently exceeding 0.9 and mAP50-95 mostly above 0.7. However, the recognition of line charts seems to be less effective, even though they achieve mAP scores above 0.7 or close to 0.7. We believe that the reason behind the relatively lower recognition scores for line charts and scatter plots might be due to their low number of pixels, making their features challenging to identify. Despite these challenges, it's worth noting that our framework achieves mAP scores above 0.7 for these categories. This demonstrates that, while there is room for improvement in recognizing line charts, our system still provides reasonably accurate and reliable results for these object types. The evaluation results show that our chart KG construction framework can achieve object recognition with high accuracy and reliability.

\subsection{Optical Character Recognition Evaluation}
\label{section7.2}
The textual content within charts primarily originates from the titles and labels of elements(Section ~\ref{section3}). We evaluate the recognition accuracy of each text category by comparing the predicted text content with the annotated text content. The specific accuracy results are depicted in the Appendix. \majorRevise{Most categories of text OCR accuracy are above 70\%. Although errors in OCR do not lead to issues in the construction of the chart structure, they can result in inaccuracies of the extracted variable names, values, and other elements within the chart, which can affect the accuracy of downstream tasks. However, in the future, OCR accuracy can be further enhanced with the introduction of more comprehensive training data and models.}

\begin{table}[ht]
\begin{center}
\caption{ 
  The results of different element recognition.
  }
\label{table4}
\renewcommand{\arraystretch}{1.3}
\begin{tblr}{cell{3}{3} = {fg=black}}

\hline
\textbf{Class}        & \textbf{Precision} & \textbf{Recall} & \textbf{mAP50} & \textbf{maP50-95} \\ \hline
\textbf{title}        & 0.999     & 0.999  & 0.990 & 0.728    \\ \hline
\textbf{x-axis title} & 0.999     & 0.999  & 0.995 & 0.828    \\ \hline
\textbf{x-axis label} & 1.000     & 0.999  & 0.995 & 0.866    \\ \hline
\textbf{y-axis title} & 0.999     & 1.000  & 0.995 & 0.908    \\ \hline
\textbf{y-axis label} & 0.999     & 1.000  & 0.995 & 0.763    \\ \hline
\textbf{legend}       & 0.997     & 1.000  & 0.995 & 0.899    \\ \hline
\textbf{bar}          & 0.999     & 0.952  & 0.962 & 0.932    \\ \hline
\textbf{line}         & 0.981     & 0.688  & 0.774 & 0.645    \\ \hline
\textbf{pie}          & 0.999     & 9.992  & 0.995 & 0.900    \\ \hline
\textbf{point}        & 0.995     & 0.986  & 0.995 & 0.876    \\ \hline
\end{tblr}
\end{center}
\end{table}

\section{Case Study}
\label{section5}
\begin{figure*}[ht]
    \begin{center}
    \includegraphics[width=\linewidth]{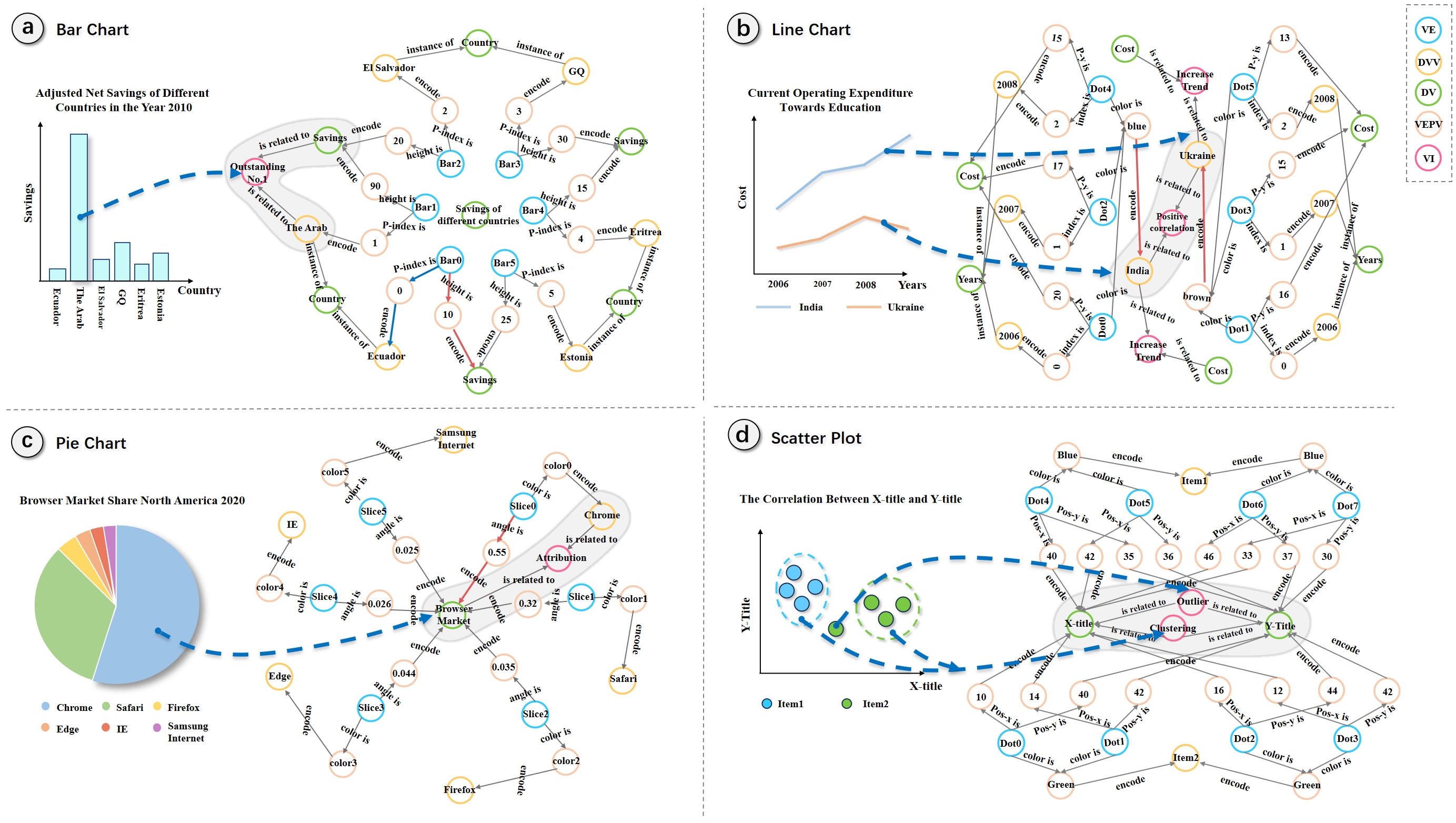}
    \caption{Examples of knowledge graph representation for four types of charts. (a) Bar Chart; (b) Line Chart; (c) Pie Chart; (d) Scatter Plot.}
    \label{fig2}
    \end{center}
\end{figure*}

To demonstrate the effectiveness of ChartKG in charts, four knowledge-graph-based representations generated for real charts and a generated chart are presented in Figure~\ref{fig2}. We will introduce the content of the charts and explicate the information captured in the knowledge graphs from three aspects: data information, visual encoding information, and visual insights.

Figure ~\ref{fig2} depicts four charts along with their corresponding knowledge graphs. The bar chart illustrates the savings of different countries in 2010, with the bar representing the Arab region notably higher than the others, making it visually prominent. The line chart displays the variations in education costs for India and Ukraine over the years. Both countries show an upward trend in education spending from 2006 to 2008, indicating a positive correlation. The pie chart illustrates the Browser Market Share in North America in 2020, where it's evident that Chrome dominates the market with a share exceeding 50\%. Lastly, the generated scatter plot exhibits two categories of data points, one of which includes noticeable outliers.

{\bf{Data information.}} In the generated knowledge graphs, two types of entities, DV and DVV, express the main topic of the chart, which indicates the semantic content of the data described in the charts. As an example, Figure ~\ref{fig2}a illustrates the adjusted net savings of different countries in the year 2010.  The information presented in the chart employs named entities, leading to their clear interpretation. Furthermore, the path VE $\rightarrow$ VEPV $\rightarrow$ DVV represents the data label bound by visual elements, while the path VE $\rightarrow$ VEPV $\rightarrow$ DV indicates the data size bound by visual elements. Specifically, the blue path in Figure ~\ref{fig2}c represents the visual element PieSlice0 with the ``Chrome'' data label, while the red path represents the data value bound by the same visual element, which is 0.55. By considering the meanings of these paths, we can infer that the proportion of Chrome is 0.55 based on the complete path meaning.

{\bf{Visual encoding.}} The chart knowledge graph showcases the utilization of visual encoding methods, which are related through VEPV $\rightarrow$ DV or VEPV $\rightarrow$ DVV. As illustrated in Figure ~\ref{fig2}a, the red path denotes the height of the bar, which represents the value of ``Savings''. Meanwhile, the position index indicates the arrangement order of elements, mapping to their respective values on the x-axis variable. In Figure ~\ref{fig2}b, the red paths represent the color encoding of lines for different countries, where blue corresponds to “India”, and brown corresponds to ``Ukraine.''

{\bf{Visual insights.}} To facilitate a comprehensive machine understanding of the underlying semantics conveyed by charts, we embed salient visual insights from the chart into the KG, as indicated by the gray shaded region in Figure ~\ref{fig2}. Specifically, as depicted in Figure ~\ref{fig2}a and c, where the ``The Ara'' bar and the ``Chrome'' pie slice are significantly higher or larger than the other elements, we represent this feature using associations between VI and DV/DVV entities. Similarly, in Figure ~\ref{fig2}b, it can be observed that there is a clear correlation between ``India'' and ``Ukraine'', with both showing a significant upward cost trend. In Figure ~\ref{fig2}d, the clustering characteristics of these points are evident between the two variables on the x-axis and y-axis, and there are also outliers present between these two variables.

\section{Example Applications}
\label{section6}
We further demonstrate the usefulness of chartKG via two example applications semantic-aware chart retrieval and visual question answering. we only need to conduct straightforward matching of graph nodes and edges for downstream chart-related tasks, which is interpretable. Firstly, we introduce a chart semantic retrieval method designed to cater to a wider range of user retrieval requirements. This highlights the flexibility and user-friendly nature of Chart KG in managing chart databases. Moreover, we accomplish the task of generating textual descriptions for charts using a predefined template approach. In comparison to existing model-based methods, our approach is lightweight and exhibits higher levels of accuracy and interpretability. \minorRevise{Additionally, since our chartKG does not focus on precise data extraction, it cannot be directly used for tasks related to data understanding. Instead, further data extraction is needed to replace the visual element property values in the KG.}

\subsection{Semantic-aware Chart Retrieval}
Chart retrieval that aligns with user preferences is a basic chart downstream task. Conventional chart retrieval methods are confined to keyword-based matches within textual elements of charts~\cite{chen2015diagramflyer}. However, with the escalation of data volume, relying solely on keywords for chart matching has become inadequate to fulfill users' retrieval needs, especially when delving into the deeper semantic expressions encapsulated by charts. For instance, the task of locating a line chart depicting a declining trend in educational expenditure from a massive pool of charts illustrates this challenge. To address this, we implement semantic-aware chart retrieval using the ChartKG. It effectively showcases the advantages of ChartKG.

\subsubsection{ChartKG-powered chart retrieval}
To enhance retrieval efficiency and facilitate user interaction, we format the user's input and employ conditional filtering. Users are prompted to sequentially provide the chart type, DV/DVV, encoding relationship, and existing visual insights. The specific steps are as follows:

Step 1: ~\majorRevise{Input formatting. Users specify the chart type, key entities, and the relationships among entities in sequence based on their desired sentence.}

Step 2: Chart type matching. As our chart dataset is stored in groups based on chart types,
it is facile to filter out a significant portion of the dataset using the specified chart type. 

Step 3: Entity matching. Leveraging our pre-extracted DV and DVV, we construct a variable dictionary to match user-input variables and variable names, further narrowing down the scope of target charts.

Step 4: Relation matching. Traversing through chart knowledge graphs that satisfy the above conditions, we sequentially match encoding relations using our defined relationship types associated with visual encodings. The remaining outcomes represent charts that align with user requirements.

\subsubsection{Evaluation}
\begin{figure*}[t]
    \begin{center}
    \includegraphics[width=\linewidth]{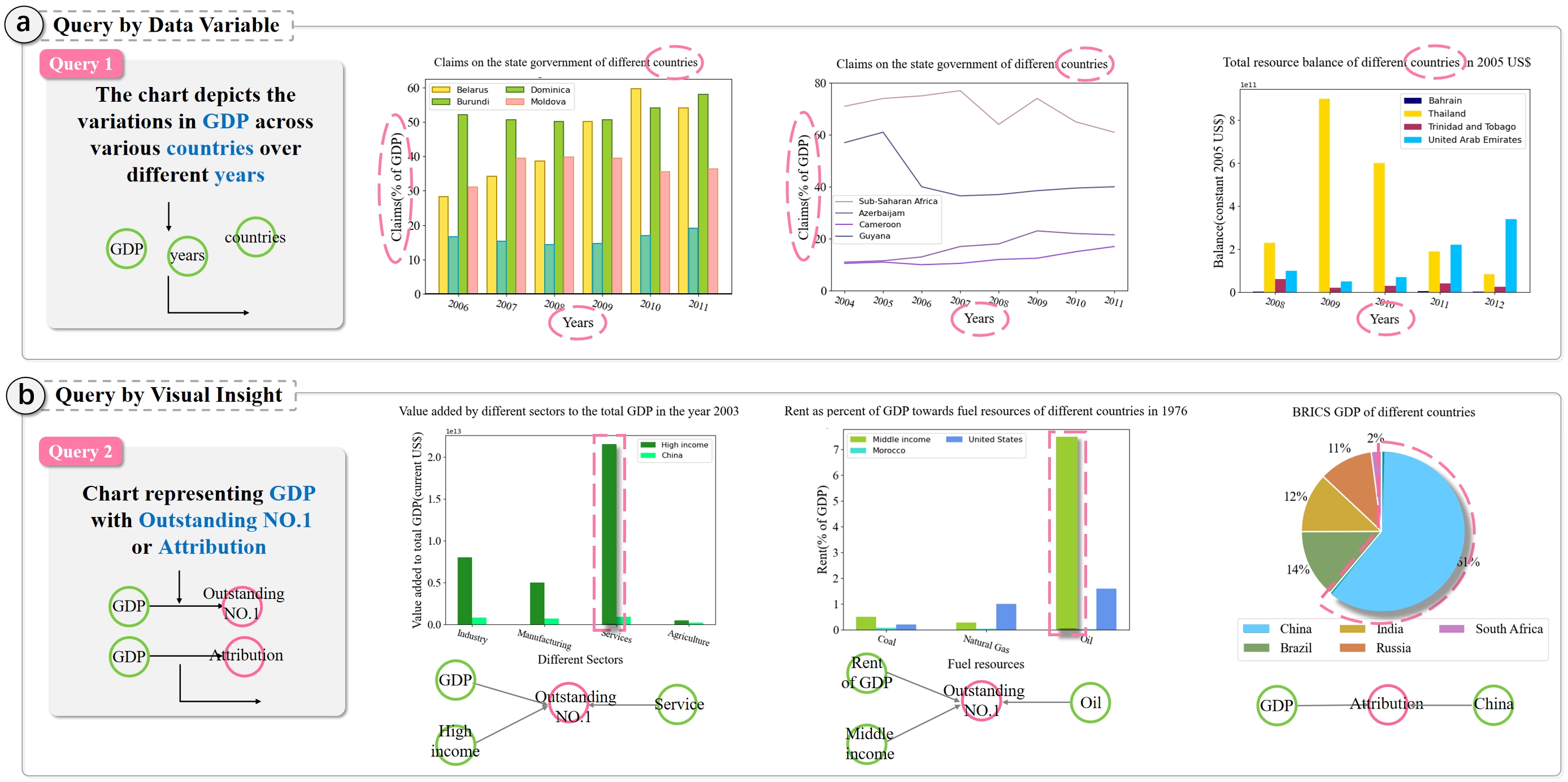}
    \caption{Example results of ChartKG-powered chart retrieval. (a) illustrates the outcomes of data variable retrieval with the keywords that meet the query requirements highlighted. (b) depicts the results of combining data variables and visual insights retrieval with the associated entities and relationships displayed below the chart.}
    \label{fig3}
    \end{center}
\end{figure*}

To demonstrate the superiority of our KG in semantic-aware chart retrieval, we conducted ten retrieval experiments, comparing with the keyword-based method ~\cite{chen2015diagramflyer} and our chartKG-powered chart retrieval method. These experiments comprised five queries targeting basic variables and another five queries targeting insights within the charts. Subsequently, we enlisted the evaluations of three domain scholars in visualization to assess the retrieval results. Some example results are presented in Figure ~\ref{fig3}. The alignment between retrieval results and retrieval criteria is crucial for evaluating the effectiveness of retrieval methods. We presented the retrieval criteria and results separately to three users, without disclosing which retrieval method corresponded to the results. We then asked them to rate the retrieval results based on whether they met the retrieval criteria, using a scale ranging from one (least satisfactory) to five (most satisfactory). The average of the ratings from the three users was taken as the final score for the retrieval results. Additionally, retrieval time is often a critical concern for users, as slow retrieval is not desirable. We also calculated the average time taken by different methods during the retrieval process to demonstrate the efficiency of our approach in terms of time.

\subsubsection{Result}
The user satisfaction with the retrieval results and the efficiency of retrieval for both methods are shown in the Appendix.
In the retrieval of variables, our chartKG-powered chart retrieval method and the keyword-based method are closely matched in terms of time and user ratings. This indicates that our ChartKG is capable of performing the most basic semantic retrieval tasks efficiently. For semantic retrieval, our average rating exceeds 4.0, which is much higher than the keyword-based method. Moreover, our time consumption is only slightly increased, by just a few seconds. This clearly demonstrates that our chartKG-powered chart retrieval method can meet users' various retrieval requirements more accurately without significantly sacrificing time efficiency.

\subsection{Visual Question Answering}
Visual question answering (VQA) is a crucial task in the field of visual data analysis~\cite{masry2022chartqa}. However, existing methods for VQA often require the joint training of both the charts and the associated questions, leading to limited scalability and interpretability of the resulting models due to the complex training process~\cite{kafle2020answering}. In contrast, utilizing a knowledge graph as a backbone can benefit VQA due to its ability to provide a structured and interpretable representation of relevant knowledge, which in turn enhances the interpretability of the VQA models and results. We demonstrate how ChartKG can enhance visual question answering. We validated its effectiveness through experiments.

\subsubsection{KG-based chart question-answering}
Through graph-based retrieval and inference, our knowledge graph enables tasks related to data comparison, visual encoding queries, and visual insight reasoning. Therefore, based on our Chart KG, we have designed three question templates, as detailed in the Appendix, to encompass aspects of data comparison, visual encoding, and visual insight. As the chart knowledge graph we proposed does not focus on the original data, specific data values are not relevant to our question design. Utilizing the chart knowledge graph and question templates, the KG-based VQA process is executed as follows: 

Step 1. Determine the question type. Initially, the user's question text is segmented into individual words, followed by determining the question type by matching keywords. 

Step 2. Locate entity relations. The knowledge graph uses word similarity matching to locate necessary entities matching the problem's words. Additionally, for data comparison inquiries, entities are grouped based on their location in the question, following keyword identification between words. 

\begin{figure*}[t]
    \begin{center}
    \includegraphics[width=\linewidth]{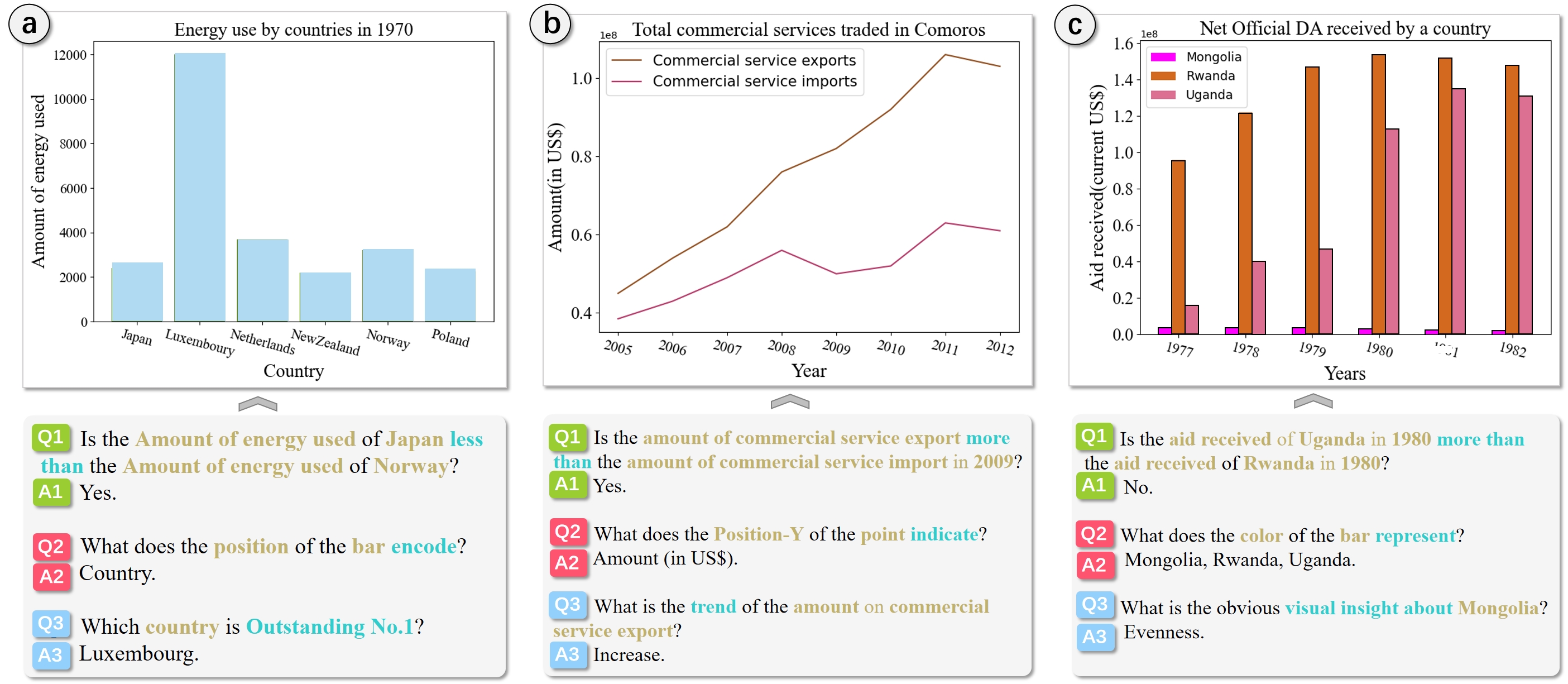}
    \caption{This figure illustrates the results of question answering for specific charts based on three question templates proposed in our work.}
    \label{fig5}
    \end{center}
\end{figure*}

Step 3. Search knowledge. Each question type corresponds to specific search rules for knowledge search, followed by inference rules for finding an answer. For example, the data value VEPV encoded by each visual element can be obtained according to the VE $\rightarrow$ VEPV$\rightarrow$ DV path, thus constituting the data sequence $X$. The grouping information of the data can be obtained based on the path VE$\rightarrow$ VEPV$\rightarrow$ DVV. If the relations between VE and VEPV represent the position index, the corresponding DVV represents the order of each visual element in the chart, resulting in the ordered sequence $Y$. Otherwise, DVV represents the category to which each visual element belongs and the group sequence $Z$ is obtained. With the sequence, $X$, $Y$, and $Z$, and whether the specified insight exists in the chart is judged based on the specific judgment mode of insight proposed in QuickInsights. We demonstrate the case of QA as shown in Figure ~\ref{fig5}.

\subsubsection{Evaluation}
To compare the superiority of our approach, we conducted a comprehensive evaluation of the effectiveness of our KG representation by calculating the average accuracy and average time spent in answering each question category, as well as overall accuracy and time.

{\bf{Model.}} T5~\cite{raffel2020exploring} is a text-to-text generation model. ChartQA~\cite{masry2022chartqa} uses a process of chart-to-table reconstruction, taking ``question, flattened table, answer" as input, and employs T5 to perform chart question-answering tasks. We chose this method as the baseline for comparison. Furthermore, to mitigate the impact of deep models on the effectiveness comparison of different representations, we substituted the flattened tables in T5 with triples from the chart KG. We named this approach KG-T5.

{\bf{Dataset.}} ChartQA provides a series of charts with corresponding tabular data. We used our question templates to generate 1 to 2 questions randomly for each chart. Additionally, we used our framework to convert charts into chart KGs as inputs for our question-answering method. Both T5 and KG-T5 were trained and evaluated on this dataset.

\begin{figure}[ht]
    \begin{center}
    \includegraphics[width=\linewidth]{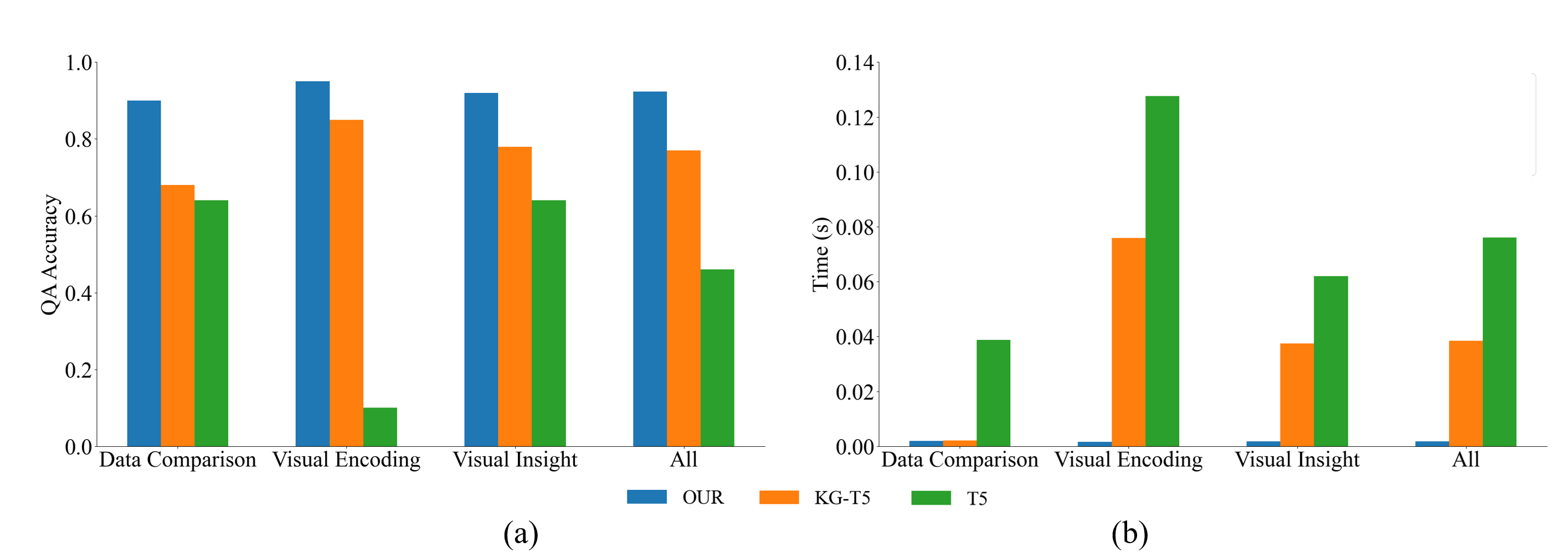}
    \caption{The evaluation of QA. (a) represents the average accuracy of the QA results, (b) represents the average time spent on QA.}
    \label{fig6}
    \end{center}
\end{figure}

{\bf{Result.}} Figure ~\ref{fig6} illustrates the performance results of the three visual QA methods. Comparing T5 and KG-T5, we found that KG-T5 achieved higher accuracy with minimal differences in time efficiency. This outcome demonstrates that our KG representation is significantly more effective in question-answering performance than relying solely on tabular data. Comparing our KG-based question-answering method to KG-T5, we observe that our method excels in both time efficiency and question-answering accuracy compared to deep learning-based methods. In terms of time efficiency, deep learning-based methods often require loading large pre-trained models, which can be time-consuming when applied to new charts. Regarding accuracy, deep learning-based chart question-answering still presents certain challenges due to its uncertainty and lack of interpretability, imposing limitations on machine understanding of charts in practical applications.

\section{Discussion}
\label{section8}
In this section, we summarize the lessons we learned during the development of Chart2KG and discuss the limitations of our approach.

\subsection{Lessons}

\subsubsection{Knowledge Graphs for Charts}
Knowledge graphs have been widely adopted for the structured representation of semantic information in natural images and are extensively applied in downstream tasks involving natural images. However, there has been limited research on constructing knowledge graphs for standard charts to model their intricate semantic relationships. Constructing knowledge graphs for charts involves two pivotal steps: defining the graph structure and constructing the graph.

{\bf{Graph Structure Definition.}} The structural definition of our graph in this paper primarily encompasses entity and relationship definitions. To ensure adaptability across various chart types, we have defined five types of entities and four types of relationships. While our graph structure provides a comprehensive representation of content in standard statistical charts, its utility in more complex visualizations may be subject to limitations. For instance, in the case of large-scale network graphs, our structure may face challenges in the application, or when the structure of the network graph itself doesn't significantly differ, the relevance of our graph structure in network graphs may not be apparent. In dealing with more intricate visualization types, a targeted approach may be necessary, such as preliminary operations like sampling and clustering for large-scale network graphs to reduce complexity. This can lead to a more meaningful application of our graph structure for representation.

{\bf{Graph Construction.}} We also present a framework for constructing the knowledge graph, which encompasses techniques such as chart classification, object recognition, and optical character recognition. In each stage, this paper adopts classical methods and conducts thorough evaluations, demonstrating the effectiveness of these conventional approaches in realizing our ChartKG. In terms of quantitative evaluation, we did not engage in direct quantitative comparisons with existing technologies but rather focused on evaluating the selected methods in isolation. While state-of-the-art techniques might perform better on these tasks, we deliberately chose more traditional methods to emphasize the simplicity and user-friendliness of our framework. Nonetheless, users maintain the flexibility to select appropriate methods based on their preferences and specific data scenarios at each stage. Furthermore, to reduce computational resource requirements, after chart categorization, we only need to train models for individual chart types during the object detection stage instead of conducting mixed training on all chart types. This approach also offers the advantage of reducing the data processing workload when incorporating new chart types. 

\minorRevise{
{\bf{Precise Data Understanding.}} The proposed \toolName{} primarily focuses on the visual semantic information of charts and does not address the extraction of raw data. We have evaluated the accuracy of data extraction using the method proposed by ChartOCR~\cite{luo2021chartocr} and the average extraction error rates for bar charts, line charts, pie charts, and scatter plots are 5.67\%, 5.65\%, 3.82\%, and 8.71\%, respectively. It is evident that extracting raw data with high precision is a challenging task. For downstream chart tasks involving specific data understanding, such as ChartQA and chart summarization, when original data is required, the extracted data can be directly used to replace the visual element property values in the chartKG. Of Course, extracting raw data can further enrich the chart KG and plays an important role in the precise analysis of chart data. We will further explore the relationships between raw data and other entities to extract a more comprehensive chart knowledge structure, providing stronger support for machine understanding of chart data.
}

\subsubsection{Comparison to KG4Vis}
In our prior work, KG4Vis~\cite{li2021kg4vis}, we established a knowledge graph for visualizations. Building upon this foundation, we present ChartKG. KG4Vis was designed for a specific task, namely, automatically recommending the construction of knowledge graphs encompassing data features, data columns, and visualization design choices. The resulting knowledge graph demonstrated effectiveness and interpretability in the context of automatic visualization recommendations. However, its applicability to other downstream tasks involving diverse chart types was limited. Motivated by this, we introduce ChartKG, which utilizes knowledge graphs to structurally represent the elements and semantics of charts themselves. Compared to bitmap images, knowledge graphs are more machine-interpretable and comprehensible. Unlike KG4Vis, ChartKG is a versatile data structure suitable for various downstream chart-related tasks, such as chart retrieval and chart question answering.

\subsection{Limitations}
\label{section LIM}
We have demonstrated the effectiveness of the chart knowledge graph through a case study and chart applications. However, there are still several limitations that need to be addressed, including the application to complex charts, potential errors in chart parsing, and evaluation challenges.

\majorRevise{
{\bf{Application to Complex Charts.}} We have conducted experiments with four types of widely used standard charts to evaluate the effectiveness of our proposed knowledge-graph(KG)-based representation and the framework for converting chart images to KG representations. \minorRevise{For more complex charts, such as arc diagrams and packed circles, our evaluation does not cover. However, for the proposed KG-based representation, our KG-based representation can still work for them by extending the current entities and relationships. 
For example, for packed circles, we need to include relationships like ``containing” or ``is the children node of” to explicitly highlight the hierarchical relationships among circles (though such hierarchical relationships can also be implicitly encoded by the center coordinate and radius of each circle). 
These complex relationships are typically between visual elements. If we can generalize and incorporate these types of relationships into the chartKG, it would better represent the knowledge of complex charts. However, different complex charts involve different customized visual elements and relationships between them, such as parent-child relationships, link relationships, and containment relationships.}
For the proposed framework of converting chart images to KG representations, it can suffer from difficulties in terms of accurately extracting and parsing the properties of visual marks and semantic information of complex charts. For example, the current framework may not work well in extracting the arcs in the arc diagrams as well as their accurate properties. But their fundamental properties, such as the starting and ending positions, can be easily extracted by the current chart-to-KG framework. Moreover, integrating advanced techniques into our framework for accurately extracting the properties of complex visual marks and semantic information can enhance its capability to handle more intricate charts.
  }

\majorRevise{
{\bf{Possible Errors in Chart Parsing.}}
At each stage of the chart-to-KG framework, we employed different technologies. We utilized ResNet for chart classification and employed object detection and OCR for chart element extraction. 
Despite their good performance, they still cannot guarantee perfect classification, detection and extraction results in our chart-to-KG framework. We analyzed the impact of error cases and potential solutions. For chart classification, incorrect classification results can lead to low accuracy in subsequent object recognition results. Object recognition selects the corresponding pre-trained model for the chart type obtained from the chart classification to perform detection. Once the chart type is misclassified, the confidence of object recognition results will generally be low. In such cases, the KG construction will be directly interrupted, and the error case will be recorded for user correction. Errors in element recognition during object recognition can lead to issues in the KG structure. For instance, when processing a bar chart, if a bar is omitted in the recognition results, the entities and relationships associated with that bar will be absent in the final KG structure. OCR errors do not lead to errors in the construction of the KG structure but manifest as incorrect names for entity nodes in the KG. This can result in reduced accuracy in downstream tasks, such as partial chart retrieval results not matching the user’s intent due to text recognition errors. 
In future work, we plan to incorporate more advanced models to reduce the error rate and develop a human-machine interaction system for manual correction of errors.
}

{\bf{Evaluation.}} While we have applied ChartKG to accomplish two chart downstream tasks, we intentionally emphasize the intrinsic effectiveness of ChartKG itself without the use of additional state-of-the-art methods for assistance. This approach lends a relatively structured and mechanized aspect to the process. For instance, we used question templates for chart-based question-answering, resulting in a somewhat uniform style in both questions and answers. Although we also employed KG as an input to Transformer-5 for question-answering, demonstrating the compatibility of our representation with deep learning models, we believe that adopting the simplest approach is more straightforward and efficient, ensuring a clearer presentation of ChartKG's core capabilities.

\section{Conclusions and Future Work}
\label{section9}
We propose ChartKG, a knowledge-graph-based representation for chart images to model the entities and relations in the chart. It consists of four types of entities and three types of relations, and can effectively represent semantic information in the visualization. To achieve the representation, we also develop chart-to-KG conversion, a framework that integrates ResNet, YOLOv5, OCR and rule-based methods to convert the bitmap-based visualization to knowledge graphs. First, Charts are classified into different types based on ResNet. Then, we extract entities based on visual element detection and entity classification and characterize chart relations through rule-based methods. We performed a quantitative evaluation to demonstrate the effectiveness of our framework, along with a case study showcasing our KG-based representation. Furthermore, based on the ChartKG, we have developed two example applications, semantic-aware chart retrieval and chart question answering. Then we conducted quantitative comparisons with several baseline methods to assess the effectiveness of our KG-based representation for chart images in downstream chart-related tasks.

In future work, we plan to integrate more types of visualization and explore knowledge mining and structured representation for infographics, as machines often struggle with comprehending infographics~\cite{Mathew_2022_WACV}. Furthermore, we aim to integrate ChartKG with textual knowledge graphs, enriching the textual knowledge repository to enhance machine comprehension of full-text documents. Additionally, considering its potential utility as prompts for large language models~\cite{shao2023prompting, guo2023images}, such integration presents an intriguing avenue for exploration.


\bibliographystyle{abbrv-doi}

\bibliography{template}


\begin{IEEEbiography}[{\includegraphics[width=1in,height=1.25in,clip,keepaspectratio]{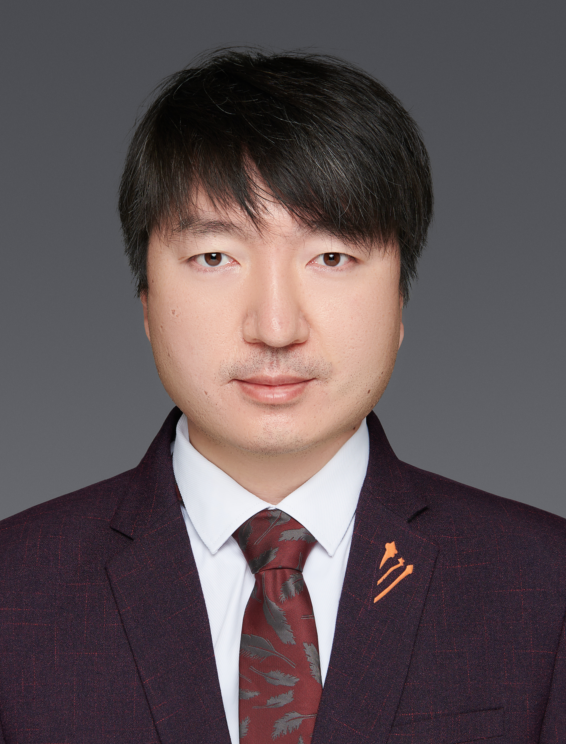}}]{Zhiguang Zhou}
is currently a professor in School of Media and Design and serves as the dean of Digital Media Technology Research Institute at Hangzhou Dianzi University. His research interests include data visualization, visual analytics and knowledge graph mining. He received his Ph.D. in Computer Science from the state key Laboratory of CAD\&CG in Zhejiang University. 
\end{IEEEbiography}

\begin{IEEEbiography}[{\includegraphics[width=1in,height=1.25in,clip,keepaspectratio]{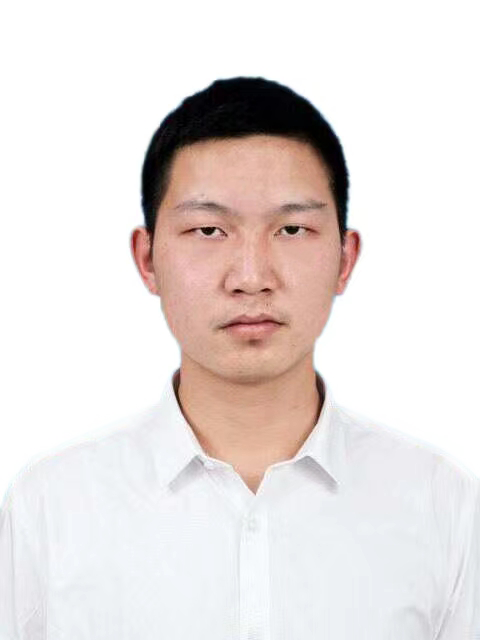}}]{Haoxuan Wang}
is currently a PhD in the School of Computer Science, Hangzhou Dianzi University, and a researcher in Big Data Visualization and Human Computer Collaborative Intelligent Laboratory. His research interests include data visualization, visual analytics and knowledge graph mining.
\end{IEEEbiography}

\begin{IEEEbiography}[{\includegraphics[width=1in,height=1.25in,clip,keepaspectratio]{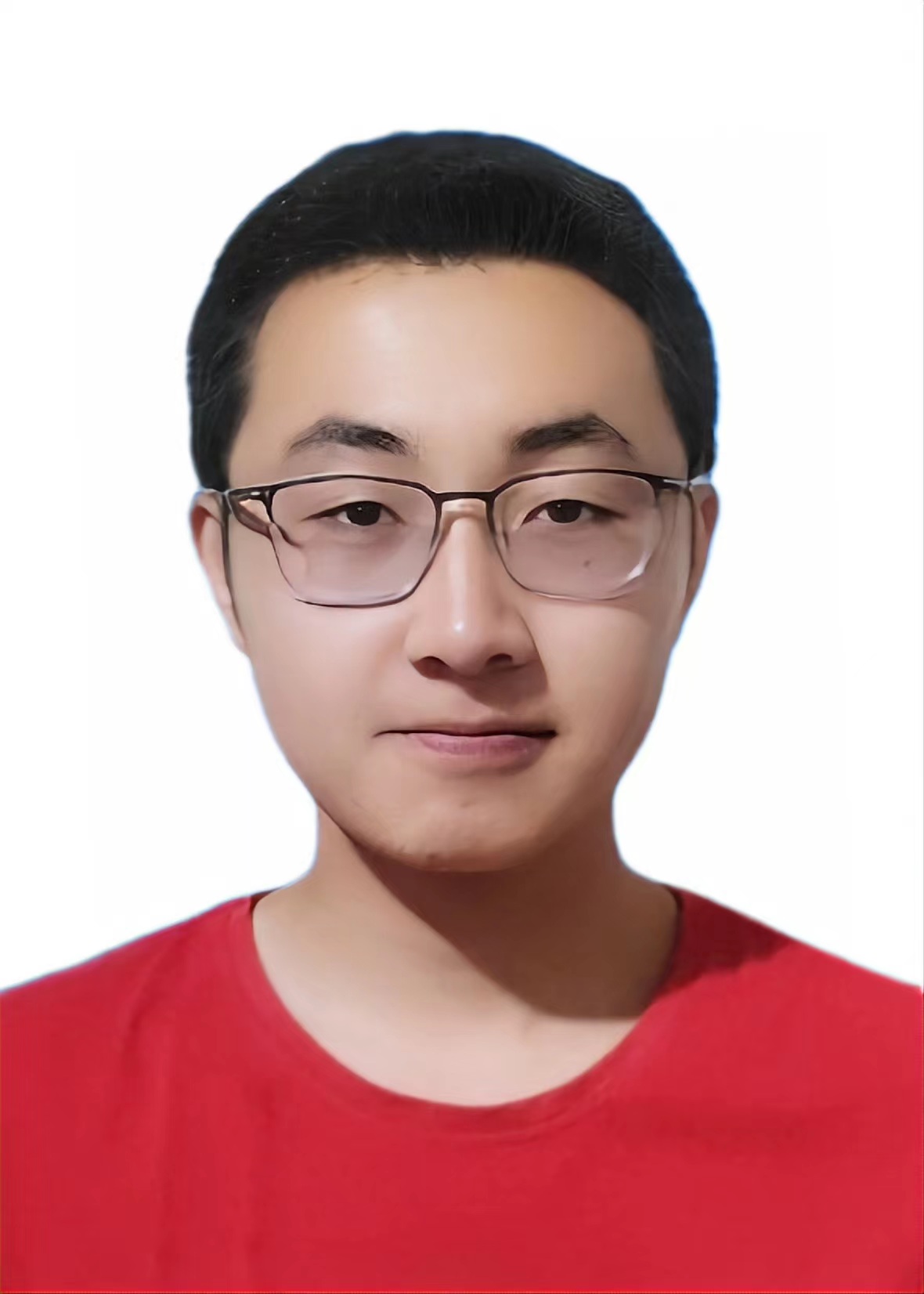}}]{Zhengqing Zhao}
is currently a Bachelor in the School of Information Management and Artificial Intelligence at Zhejiang University of Finance and Economics and a researcher in Big Data Visualization and Human Computer Collaborative Intelligent Laboratory. His research interests include data visualization and knowledge graph mining.
\end{IEEEbiography}

\begin{IEEEbiography}[{\includegraphics[width=1in,height=1.25in,clip,keepaspectratio]{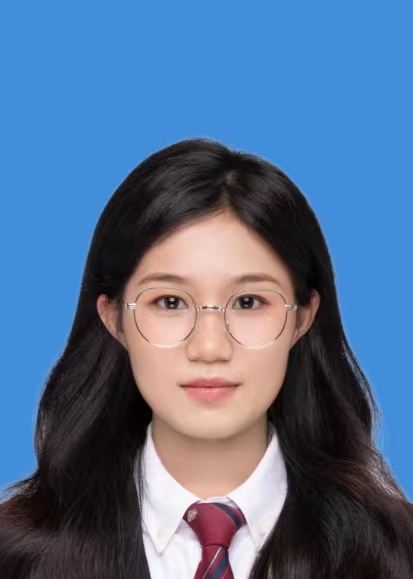}}]{Fengling Zheng}
is currently a PhD in the School of Computer Science, Hangzhou Dianzi University, and a researcher in Big Data Visualization and Human Computer Collaborative Intelligent Laboratory. Her research interests include interactive visualization, visual analytics and knowledge graph.
\end{IEEEbiography}

\begin{IEEEbiography}[{\includegraphics[width=1.0in,height=1.25in,clip,keepaspectratio]{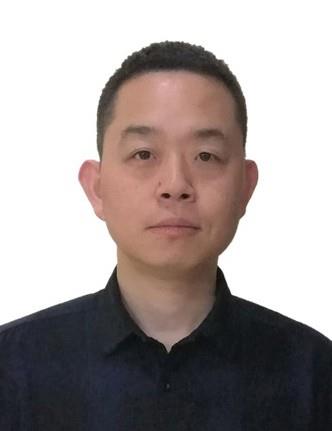}}]{Yongheng Wang} received the Ph.D. degree in computer science and technology from the National University of Defense Technology, Changsha, China, in 2006. He is currently a Research Specialist with the Research Center of Big Data Intelligence, Zhejiang Lab, Hangzhou, China. His research interest covers big data analysis, machine learning, computer simulation, and intelligent decision making.
\end{IEEEbiography}

\begin{IEEEbiography}[{\includegraphics[width=1.0in,height=1.25in,clip,keepaspectratio]{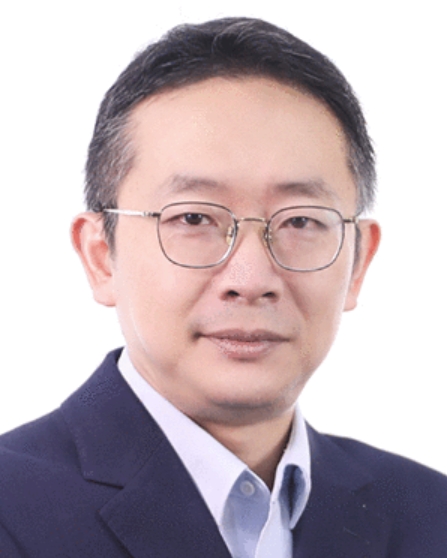}}]{Wei Chen} is a professor with the State Key Lab of CAD\&CG, Zhejiang University. His research interests include visualization and visual analysis. He has published more than 70 IEEE/ACM Transactions and IEEE VIS papers. He actively served as guest or associate editors of the ACM Transactions on Intelligent System and Technology, IEEE Computer Graphics and Applications and Journal of Visualization.
\end{IEEEbiography}

\begin{IEEEbiography}[{\includegraphics[width=1.0in,height=1.25in,clip,keepaspectratio]{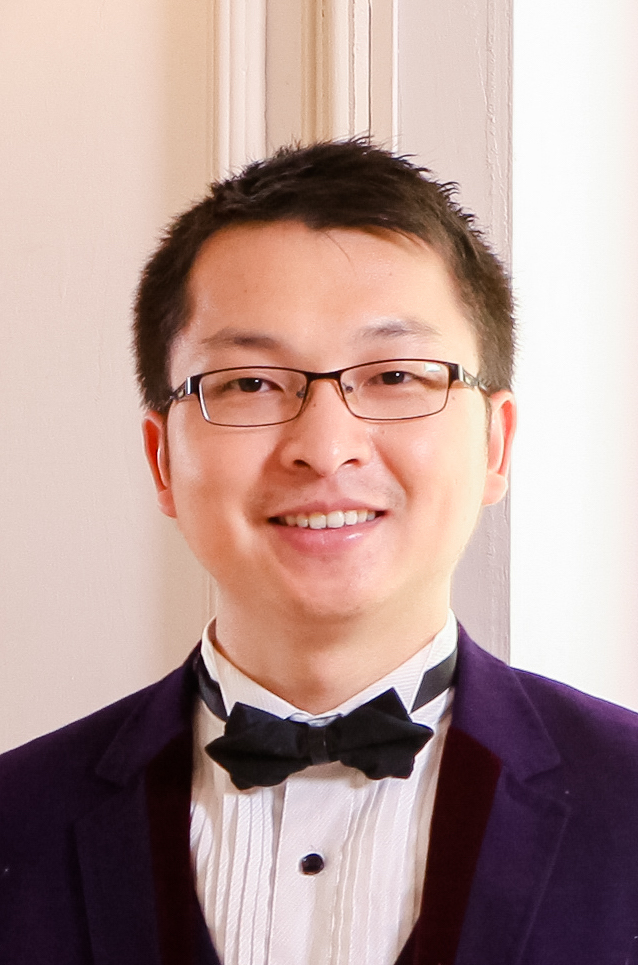}}]{Yong Wang}
is currently an assistant professor in the College of Computing and Data Science, Nanyang Technological University. Before that, he worked as an assistant professor at Singapore Management University from 2020 to 2024. His research interests include information visualization, visual analytics and human-AI collaboration, with an emphasis on their application to FinTech, quantum computing and online learning. He obtained his Ph.D. in Computer Science from Hong Kong University of Science and Technology. He received his B.E. and M.E. from Harbin Institute of Technology and Huazhong University of Science and Technology, respectively. For more details, please refer to \url{http://yong-wang.org}.

\end{IEEEbiography}


\end{document}